\documentclass[letterpaper, 10 pt, conference]{ieeeconf}  
\usepackage{amsmath,amsfonts}
\usepackage{mathrsfs}
\usepackage{textcomp}
\usepackage{stfloats}
\usepackage{url}
\usepackage{cite}
\usepackage{amssymb}
\usepackage[dvipsnames]{xcolor}
\usepackage{balance}
\usepackage{graphicx}
\usepackage{color}
\usepackage{svg}
\usepackage{xfakebold}

\usepackage{booktabs}
\usepackage{array}
\newcommand{\rotcol}[1]{\rotatebox{25}{\parbox{2.3cm}{ #1}}}
\usepackage[font=footnotesize]{caption}
\usepackage{subcaption}

\newcommand{\fbseries}{\unskip\setBold\aftergroup\unsetBold\aftergroup\ignorespaces}
\newcommand{\setBoldness}[1]{\def\fake@bold{#1}}
\newcommand{\hlcode}[1]{%
  {\transparent{0.12}\colorbox{box}{\transparent{1.0}#1}}%
}
\let\labelindent\relax
\usepackage{enumitem}
\usepackage{listings}

\definecolor{codegreen}{rgb}{0,0.6,0}
\definecolor{codegray}{rgb}{0.8,0.8,0.8}
\definecolor{commentgrey}{rgb}{0.5,0.5,0.5}
\definecolor{backcolour}{rgb}{0.98,0.98,0.98}
\definecolor{functionblue}{rgb}{0.1,0.1,0.7}
\definecolor{box}{HTML}{F48A00}

\lstdefinestyle{myPython}{
    backgroundcolor=\color{backcolour},   
    commentstyle=\color{commentgrey},
    keywordstyle=\color{codegreen},
    numberstyle=\tiny\color{codegray},
    stringstyle=\color{commentgrey},
    basicstyle=\ttfamily\footnotesize\fbseries,
    morekeywords={as},
    breakatwhitespace=false,         
    breaklines=true,                 
    captionpos=b,
    emph={policy, policy_v0, policy_v1, policy_v2, evaluate, objective_fn},
    emphstyle={\color{functionblue}},
    keepspaces=true,                 
    numbers=none,                    
    numbersep=5pt,                  
    showspaces=false,                
    showstringspaces=false,
    showtabs=false,                  
    tabsize=2
}

\IEEEoverridecommandlockouts                              

\makeatletter
\let\NAT@parse\undefined
\makeatother
\usepackage{hyperref}
\hypersetup{
  colorlinks=true,
  linkcolor=red,
  urlcolor=blue,
  citecolor=green,
}
\title{\LARGE \bf
Combining Large Language Models and Gradient-Free Optimization for Automatic Control Policy Synthesis}
\author{Carlo Bosio$^1$, Matteo Guarrera$^2$, Alberto Sangiovanni-Vincentelli$^2$, Mark W. Mueller$^1$
\thanks{UC Berkeley, Department of Mechanical Engineering$^1$ and Department of Electrical Engineering and Computer Sciences$^2$,
Contact:{\tt\small \{c.bosio, matteogu, alberto, mwm\}@berkeley.edu}}
}
\begin{document}
\maketitle
\thispagestyle{empty}
\pagestyle{empty}
%



\begin{abstract}
Large Language models (LLMs) have shown promise as generators of symbolic control policies, producing interpretable program-like representations through iterative search. However, these models are not capable of separating the functional structure of a policy from the numerical values it is parametrized by, thus making the search process slow and inefficient. We propose a hybrid approach that decouples structural synthesis from parameter optimization by introducing an additional optimization layer for local parameter search. In our method, the numerical parameters of LLM-generated programs are extracted and optimized numerically to maximize task performance. With this integration, an LLM iterates over the functional structure of programs, while a separate optimization loop is used to find a locally optimal set of parameters accompanying candidate programs.
We evaluate our method on a set of control tasks, showing that it achieves higher returns and improved sample efficiency compared to purely LLM-guided search. We show that combining symbolic program synthesis with numerical optimization yields interpretable yet high-performing policies, bridging the gap between language-model-guided design and classical control tuning. Our code is available at
\url{https://sites.google.com/berkeley.edu/colmo}.
\end{abstract}
%
%
\section{Introduction}
Designing control policies that are both high-performing and interpretable remains a central challenge in robotics and control\cite{doshi2017towards}. Classical control methods such as PID or optimal control provide guarantees, but they typically require expert modeling and significant manual effort. At the same time, modern machine learning techniques—most notably deep reinforcement learning (RL)—can discover highly performant policies from data, but often produce opaque neural networks that are difficult to interpret, verify, or deploy in safety-critical systems.

\begin{figure}[!tb]
    \centering
    \includegraphics[width=0.9\linewidth]{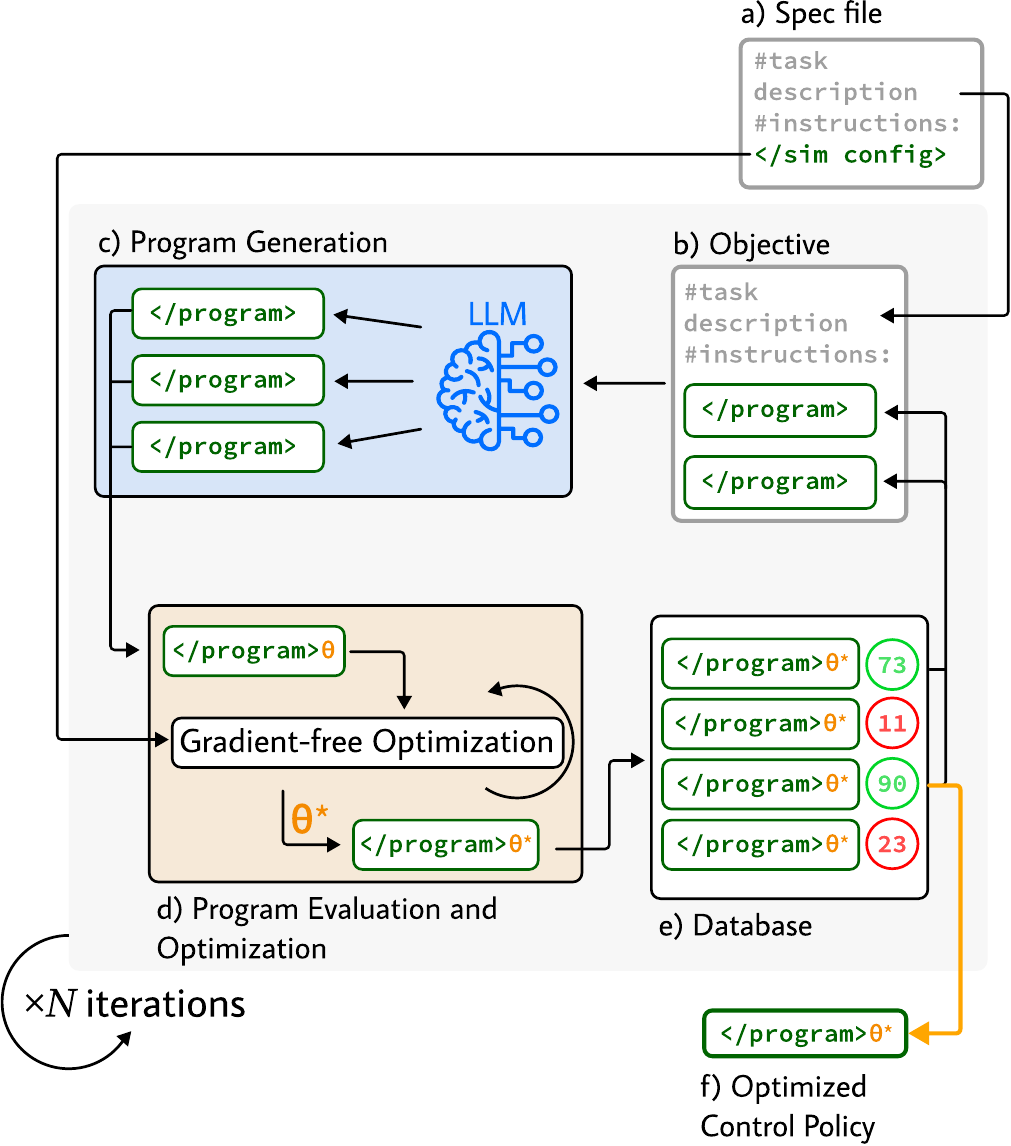}
    \caption{Block diagram of the algorithmic infrastructure for our policy search method. The input to the algorithm is a specification file a) containing a task description, the implementation of an evaluation function to score programs, and some starter code to evolve. A prompt b) is constructed pasting the current best programs (the starter code at the beginning). The prompt is fed to a \textit{Program Generation} block c) containing a pre-trained LLM, which produces more programs. The control policies generated by the LLM are fed to the \textit{Program Evaluation and Optimization} block d), which optimizes them and scores them based on their closed-loop performance in simulation. The program-score pairs are stored in a \textit{Database} e), from which they are sampled to be included in following prompts and improved. The output is a high-performance, programmatic control policy f).}
    \label{fig:block-diag}
    \vspace{-0.3cm}
\end{figure}

Recent work has explored the use of LLMs as generators of symbolic control policies, framing the problem as a form of program synthesis guided by reinforcement feedback. By iteratively generating and evaluating candidate programs in a genetic algorithm framework, it is possible to produce compact, human-readable controllers
\cite{bosio2025synthesizing}.~This line of work provides a computational tool to automatically synthesize interpretable and modular control strategies. 

However, a key limitation of current LLM-driven approaches is their reliance on the LLM to simultaneously specify both structure and numerical parameters (e.g., control gains, thresholds). LLMs are good at producing syntactically correct and qualitatively meaningful code, but they are not well-suited for choosing precise numerical values. As a result, many generated policies that would perform well if associated to the right set of parameters, in reality perform poorly until enough evolutionary steps stumble upon suitable values—an inefficient and potentially brittle process.

With our hybrid symbolic-numeric search, we propose to decouple program structure generation from parameter tuning by introducing a numerical optimization stage in the loop with the program search process. In our approach (see Fig.~\ref{fig:block-diag}), the LLM-generated programs are processed to replace numerical parameters with symbolic placeholders, and are then optimized using a zeroth-order numerical optimizer. This hybrid approach allows:
\begin{itemize}[leftmargin=*]
    \item The LLM to generate functional structure of candidate controllers.
    \item The optimizer to exploit gradient-free search to find a locally optimal set of parameters.
\end{itemize}

We show that this combination leads to higher performances and improved sample efficiency (i.e., number of search iterations to reach a target performance) compared to pure LLM-driven search. Our results suggest that separating symbolic reasoning from numeric search is a powerful paradigm for automatic policy synthesis and could accelerate the adoption of programmatic controllers in real-world applications. 

The contributions of this work are as follows:
\begin{itemize}[leftmargin=*]
    \item Proposing a novel symbolic-numeric method for automatic synthesis of programmatic control policies;
    \item Achieving over one order of magnitude fewer search steps in compared to purely LLM-driven search to achieve target performances;
    \item Conducting extensive testing on control tasks, including full-scale quadruped locomotion;
    \item Open-sourcing our framework based on low compute budget requirements.
\end{itemize}


%
\section{Related Work\label{sec:related-work}}
In the following we present some of the previous work which motivated the development of this idea.
\subsection{Program synthesis and Large Language Models\label{sec:prog-synth}}
Program synthesis \cite{gulwani2017program} is the task of automatically finding a program expressed in some underlying programming language and satisfying user intents and constraints. These methods have been limited by the large dimensions of the search space \cite{gulwani2010dimensions}. To make the search more tractable, a common strategy is to use manually designed Domain Specific Languages (DSLs) \cite{cui2024reward, patton2023program}, which are application-specific. Another approach is the use of sketches, where partial program structures are provided by the user and then completed by the synthesis algorithm \cite{solar2009sketching}. 
LLMs for code \cite{fried2022incoder, li2023starcoder, lozhkov2024starcoder}, have been used as a DSL-free alternative to program synthesis.  
On top of LLMs for code generation, extensive system-level research has been proposed with the aim of integrating LLMs with additional components to achieve more complex tasks than with single prompt engineering. A popular example is the combination of LLMs with evaluators, i.e. programmatic ways of scoring their outputs, and evolution strategies such as genetic algorithms \cite{romera2024mathematical}. This paradigm has been applied in various contexts, such as automated reasoning \cite{zelikman2022star}, code debugging \cite{haluptzok2022language}, and algorithm design \cite{lehman2023evolution, liu2024example}.

\subsection{Interpretable and Programmatic Reinforcement Learning}
There has been increasing interest within the machine learning community in intersections between program synthesis and data-driven methods \cite{balog2016deepcoder}. In fact, programmatic and symbolic representations improve the interpretability of learned functions and policies. For example of these works are learning verifiable decision-tree policies \cite{bastani2018verifiable}, and programmatic RL \cite{verma2018programmatically, verma2019imitation} where policies are synthesized from a domain-specific language. These methods offer interpretability but require carefully designed DSLs or templates, limiting expressivity. A DSL-free approach to control policy synthesis using LLMs was proposed in \cite{bosio2025synthesizing}, where a genetic search framework is used to iteratively synthesize controllers. However, all these LLM-based approaches bypass the numerical aspect of control design, relying on the discrete program search to guess suitable parameter values. This leads to inefficient exploration of the continuous parameter space and often yields suboptimal performance. Our method, by incorporating in-the-loop numerical optimization, introduces a systematic local search in the parameter space, complementing program synthesis.

\subsection{Gradient-free Optimization in Control Design}
%
Gradient-free optimization (GFO) \cite{conn2009introduction} is a family of methods for function optimization in cases where derivatives are unavailable or unreliable.
In control design, GFO methods have been widely applied to tasks such as auto-tuning of PID gains or model predictive control cost weights \cite{neumann2019data, berkenkamp2023bayesian, sartore2024automatic} in simulation-in-the-loop or hardware-in-the-loop settings.
Such optimizers typically require a fixed control structure, and thus cannot discover fundamentally new strategies. Our work integrates these optimizers into a structure-search loop, enabling them to tune parameters for policy candidates automatically generated by an LLM-driven search framework. Our work draws inspiration from GFO-based control tuning to find local optima in the parameter space for the LLM-generated policies.

\subsection{Hybrid Symbolic–Numeric Methods}
Our work is related to recent efforts combining symbolic reasoning with numerical search. For example, symbolic regression systems (e.g., \cite{cranmer2023interpretable, udrescu2020ai}) discover analytic expressions from data and then optimize their parameters. In the context of LLMs, recent studies have explored LLMs generating differentiable code for physics simulation that is optimized via gradient-based methods \cite{ma2024llm}. 
Related work has also employed LLMs for automatic design of electronic circuits \cite{lai2025analogcoder}. We extend this symbolic-numerical paradigm to control synthesis. In this work, we bring these symbolic-numeric ideas to develop an automatic synthesis tool for programmatic controllers.


\section{Methodology \label{sec:method}}
%
Our approach enables developing interpretable, high-performance policies for the control tasks of interest.
We first briefly describe the problem fundamentals, following the formalism of \cite{bertsekas2019reinforcement}, and then the algorithmic aspects of our approach.

\subsection{Background}
We wish to find a control policy for a discrete-time dynamical system with dynamics in the form
\begin{equation}
    x_{t+1} = f(x_t, u_t),
\end{equation}
where $t\in\mathbb{N}$ is the time index, $x_t\in\mathbb{R}^n$ is the state of the system at time step $t$, and $u_t\in U \subset \mathbb{R}^m$ is the control input. At each time step $t$, a stage reward
\begin{equation}
    r_t = g(x_t, u_t)\label{eq:step-reward}
\end{equation}
is incurred. The general objective is to find a control policy $u_t = h(x_t)$ to maximize the cumulative reward 
\begin{equation}
    R = \sum_{t=0}^T r_t, \label{eq:return}
\end{equation} 
where $T$ is a specified time horizon. 
In \cite{bosio2025synthesizing}, the function $h(\cdot)$ is represented as a non-parametric program $\texttt{policy}(\cdot)$ in Python, i.e.
\begin{equation}
    u_t = \texttt{policy}(x_t).
\end{equation}
The goal is then to find a controller $\texttt{policy}^*(\cdot)$ by approximating the solution to the following optimization problem:
\begin{equation}
\begin{aligned}
    \texttt{policy}^*(\cdot)= & \max_{\texttt{policy}(\cdot)} \,\,\, R &&&\\
     & \,\,\text{s.t.} \,\,\,\, x_{t+1} = f(x_t, u_t), \forall t\in\{0, ..., T\}&&\\
     & \,\,\,\,\,\,\,\,\,\,\,\,\, u_t = \texttt{policy}(x_t),&& 
\end{aligned}
\label{eq:opt-problem}
\end{equation}
where $R$ is defined in eq. \ref{eq:return}.
The search in this case does not happen in a parameter space encoding a mathematical structure (such as, for instance, a neural network), but directly in the space of programs, which is infinite-dimensional and complex to search over for the reasons discussed in Section~\ref{sec:prog-synth}. An LLM is used in \cite{bosio2025synthesizing} to guide the program search. 
\subsection{Parametric program approach}
The use of an LLM shifts the optimization to the space of tokens, i.e. the elements LLMs use to decompose a string of text in elementary units. 
This formalization leads to some results in simple settings. However, it does not provide the numerical stability and robustness that are needed in control, where a small deviation in parameters can lead to numerical instabilities.
Furthermore, LLMs have notoriously struggled with simple math tasks and multi-step numerical reasoning, often producing incorrect results. 
Additionally, LLMs are not designed to operate on floating point numbers, which can make this vanilla search approach brittle. These limitations motivate separating structural synthesis from numerical optimization.

We propose an improvement of the formulation~\eqref{eq:opt-problem}, in which the control program is parametric, i.e.
\begin{equation}
    u_t = \texttt{policy}(\theta; x_t),\label{eq:param-prog}
\end{equation}
where the parameters $\theta$ are all the continuous parameters that can be found in a controller, such as control gains, weights, thresholds. The values produced by an LLM are likely to be suboptimal, or not tuned for the specific task of interest. We alleviate this problem by adding an inner GFO loop optimizing the parameters $\theta$:

\begin{equation}
\begin{aligned}
    \texttt{policy}^*(\cdot)= & \max_{\texttt{policy}(\cdot) \in \Phi} \,\,\,\, \max_{\theta} \,\,\, R &&&\\
     & \,\,\,\,\,\,\,\,\,\,\,\,\,\,\,\,\,\,\,\text{s.t.} \,\,\,\, x_{t+1} = f(x_t, u_t), \forall t\in\{0, ..., T\}&&&\\
     &\,\,\,\,\,\,\,\,\,\,\,\,\,\,\,\,\,\,\,\,\,\,\,\,\,\,\,\,\,\,\, u_t = \texttt{policy}(\theta; x_t),&&& 
\end{aligned}
\label{eq:opt-problem-opt}
\end{equation}

Compared to \eqref{eq:opt-problem}, the new formulation \eqref{eq:opt-problem-opt} consists of a nested optimization problem. The outer layer optimizes over the functional structure of programs, while the inner layer finetunes the programs finding the optimal numerical parameters. This enables to reduce the number of LLM-generated samples required to find high-performing policies.
The practical realization of these high-level ideas is inherently nontrivial, owing to several challenges that we detail in the following sections.


\section{Implementation details}
The overall algorithmic infrastructure is shown in Fig.~\ref{fig:block-diag}.
The input is a specification file, where a description of the task, some starter code for the function to evolve and the implementation of the evaluation function are provided. At each iteration, a \textit{Program Generation} block (containing the LLM) produces candidate control programs for the task of interest. The proposed control policies are then fed to a \textit{Program Optimization} block, which simulates the underlying system in closed loop and optimizes the candidate program's parameters. The best performing programs are stored in a \textit{Database} and added to the prompts for the subsequent iterations, where the LLM is instructed to improve upon the previously generated programs. In the following we explain in more detail the functioning of this framework. In the following sections, we briefly describe the algorithm components, with particular emphasis on the novel parts, and we refer to \cite{bosio2025synthesizing} for additional details on the base infrastructure.
%
\subsection{Specification and Prompt Construction}
The input to our control synthesis framework is a specification file (Fig. \ref{fig:block-diag}a), composed by three main parts:
\begin{itemize}
    \item A description of the control task to solve;
    \item An initial candidate control program in the form of starter code;
    \item The evaluation function used by the program optimization block (Fig. \ref{fig:block-diag}d).
\end{itemize}
An example of a general specification structure for our control synthesis method is shown in Fig.~\ref{fig:spec}. 

At each iteration, a prompt (Fig. \ref{fig:block-diag}b) is constructed by concatenating two previously generated high-performing programs to the task description from the specification file (see Fig. \ref{fig:prompt}). The LLM is thus instructed to improve upon the two provided programs.
During the initial phases of the algorithm execution, only the starter code (from the specification file) is available, and is directly pasted into the prompt. 
As the evolution progresses and higher-performing programs are generated, these  are used in the prompt instead of the starter code.
An example of a general prompt structure is shown in Fig. \ref{fig:prompt}.
%
\begin{figure}[tb]
    \centering
    \begin{lstlisting}[language=Python, escapechar=!]
 """Finds a policy for the control task."""

 # import libraries needed
 import numpy as np

 # evaluation function for the control policy
 def evaluate(params):
   """Returns optimized params and score."""
   !\transparent{0.12}\colorbox{box}{\transparent{1.0}opt\_params, score = Optimize(objective\_fn)}!
   return opt_params, score

 def objective_fn(params):
   """Returns the simulation reward."""
   environment = initialize_env()
   obs = environment.get_observation()
   total_reward = 0.0
   for _ in range(1000):
     action = policy(params, obs)
     reward, obs = environment.step(action)
     total_reward += reward
   return total_reward

 # function to evolve
 def policy(params, observation):
   """Returns a control action."""
   action = np.random.uniform()
   return action
\end{lstlisting}
    \caption{Example pseudo-code template for a control synthesis specification. The \texttt{objective\_fn} function is wrapped into a GFO loop (in orange).}
    \label{fig:spec}
    \vspace{-0.3cm}
\end{figure}
\subsection{Program Generation} 
A pre-trained LLM is the generation engine of the \textit{Program Generation} block (Fig. \ref{fig:block-diag}c), in which candidate control programs are sampled.
The generation step from the provided prompt can be thought of as a genetic operator \cite{mitchell1998introduction}, since, among all the possible mechanisms that describe an LLM output based on its inputs, these possibilities can occur:
\begin{itemize}[leftmargin=*]
    \item The two programs provided in the prompt are mixed (i.e., \textit{crossover});
    \item One of the programs provided is modified (i.e., \textit{random mutation});
    \item An entirely new program is produced.
\end{itemize}
%
%
\subsection{Program Evaluation}
The evaluation block (Fig. \ref{fig:block-diag}d) is the core innovation of our method. 
Candidate control programs are automatically extracted from the LLM output and undergo string-level preprocessing operations. The numerical parameters in a program are likely not very accurate, since they are LLM-generated. To alleviate this, all the numerical values are parsed through a Regular Expression (RegEx) \cite{friedl2006mastering} and replaced with the substrings \texttt{params[0]}, \texttt{params[1]}, \texttt{params[2]}, etc. with progressive indices, and the policy signature is modified so to accept the \texttt{params} variable as an input. These operations make the program parametric. 
The program parametrization is then fed to the GFO optimizer, which iteratively evaluates the candidate program with refined sets of parameters. 
At each GFO iteration, the candidate controllers are simulated in closed loop with the task of interest. The performance is quantified by the reward (\ref{eq:return}) provided in the specification file. 
Syntactically incorrect or incomplete programs are discarded, while promising program-score pairs are stored in the \textit{programs database}, from which they are sampled to be added to the subsequent prompts, and evolved.
We found that in most cases the dimension of the parameter vector is less than 20, which falls within a suitable regime for GFO algorithms \cite{conn2009introduction}. 

%
\subsection{Programs Database}
The generated high-performing programs are stored in the \textit{programs database} (Fig. \ref{fig:block-diag}e). 
Each program is assigned a unique identifier derived solely from its symbolic representation to make sure that programs differing only in their parameters but sharing the same functional structure are not redundantly stored.
In this way, we avoid feeding back to the LLM two programs that are structurally equivalent.
To discourage getting stuck in local optima, an island approach is implemented, where different instances of the program search are run independently \cite{cantu1998survey}. 
Thus, when constructing a prompt, a two stage sampling procedure happens: first, an island is sampled, then programs contained within the selected island are sampled to be added to the prompt. More details on the database implementation can be found in \cite{bosio2025synthesizing} and \cite{romera2024mathematical}.
%
\begin{figure}[tb]
    \centering
    \begin{lstlisting}[language=Python]
 """Finds a policy for the control task.
    On every iteration, improve policy_v1 over the policy_vX methods from previous iterations.
 """
 import numpy as np

 def policy_v0(observation):
   """Returns a control action."""
   action = np.random.uniform()
   return action
   
 def policy_v1(observation):
   """Returns a control action."""
   action = 0.0
   return action

 def policy_v2(observation):
   """ Improved version of policy_v1."""

\end{lstlisting}
    \caption{Example pseudo-code template for a prompt. The LLM generates a body for the provided function signature trying to improve upon previously generated functions. \texttt{policy\_v0} and \texttt{policy\_v1} are sampled from the database, the best performing policies are sampled more frequently.}
    \label{fig:prompt}
    \vspace{-0.3cm}
\end{figure}
\subsection{Computational Considerations}
\begin{figure}[tb]
    \centering
    \includegraphics[width=0.7\linewidth]{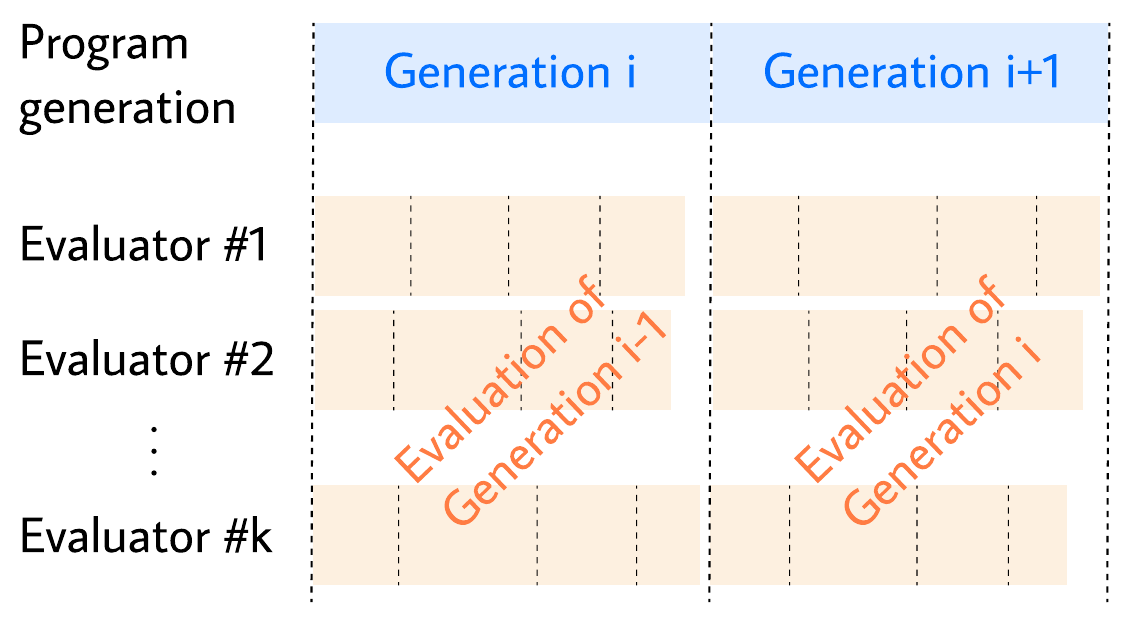}
    \caption{Parallel implementation of program generation and evaluation. The program generation (in blue) happens on the GPUs and evaluation (in orange) happens on the CPUs. In this way, the computational cost of optimization is completely shadowed.}
    \label{fig:threads}
\end{figure}

A number of hyperparameters affect the generation performances of an LLM \cite{bosio2025synthesizing}. 
As a general rule of thumb, we found that a larger batch size (i.e., the number of output programs generated in parallel, from the same prompt) is always better, since it leads to more exploration. The LLM inference time scales sublinearly with the batch size thanks to GPU parallel inference, therefore it is always recommended to work with the largest batch sizes allowed by the available hardware. 
Adding a GFO optimization in the loop may sound like an increase in computational cost. However, given the significant time required for the LLM inference (in the order of tens of seconds), the GFO evaluation can be shadowed through multithreading: during LLM inference (on GPUs), the evaluation block running on CPUs evaluates and optimizes the previously generated program batch (see Fig. \ref{fig:threads}). This overlapping utilization ensures that no resource remains underused. As long as the evaluation is faster than the generation, the queue of programs remains bounded. Depending on the specific GFO algorithm used, the evaluation time can vary. We tested Bayesian Optimization (BO) \cite{frazier2018tutorial} and Evolutionary Strategies (ES) \cite{hansen2015evolution}. BO is more sample efficient but slower, while ES approaches require more iterations to converge but are computationally cheaper. For our use case, we obtained the best results with ES.
%
\section{Results\label{sec:casestudies}}
In the following we provide more detail about the practical aspects of the implementation of our method. We then present the results obtained on a set of simulated control tasks.
\subsection{Setup}
All experiments were conducted using two NVIDIA GeForce RTX 3090 GPUs and a single AMD Ryzen Threadripper 3990X 64-Core Processor. The model used in the program generation block (Fig. \ref{fig:block-diag}c) is the 7-billion-parameter model from the Qwen 2.5 family of models for code\cite{hui2024qwen2}. To fully take advantage of the available hardware and maximize the batch sizes per LLM inference, we used 4-bits weight quantization\cite{dettmers2022llmint88bitmatrixmultiplication} and model pipelining \cite{huang2019gpipe}. All GFO algorithm implementations were taken off-the-shelf from the open-source library Nevergrad \cite{bennet2021nevergrad}. The results we show are obtained running 100 iterations of the (1+1)-ES algorithm~\cite{hansen2015evolution} in the GFO block (Fig. \ref{fig:block-diag}d). All tasks are simulated through the open-source simulator MuJoCo \cite{todorov2012mujoco}, either directly or through the Control Suite \cite{tassa2018deepmind} library.


\begin{table}[tb]
\centering
\caption{State and input space dimensions for each tasks. Performances of the synthesized policies for each task using only LLM-driven search and adding the in-the-loop GFO.}
\vspace{-0.3cm}
\begin{tabular}{>{\arraybackslash}m{2.3cm}*{4}{>{\arraybackslash}m{0.7cm}}}
               & \rotcol{State dim.} & \rotcol{Action dim.} & \rotcol{Reward (no GFO)} & \rotcol{Reward (GFO)}\\ \midrule
Pendulum swing-up & 2 & 1 & 533 & \textbf{591} \\ \midrule
Ball in cup    & 8 & 2 & 575 & \textbf{755} \\ \midrule
Cheetah        & 18 & 6 & 159.1 & \textbf{252} \\ \midrule
Quadruped      & 36 & 12 & 62.1 & \textbf{181.3} \\ \midrule
Unitree A1       & 36 & 12 & 33.8 & \textbf{152.5} \\ \bottomrule
\end{tabular}
\label{tab:task-dimensions}
\vspace{-0.2cm}
\end{table}
\subsection{Case Studies}
\begin{figure}[tb]
    \centering
    \includegraphics[width=0.99\linewidth]{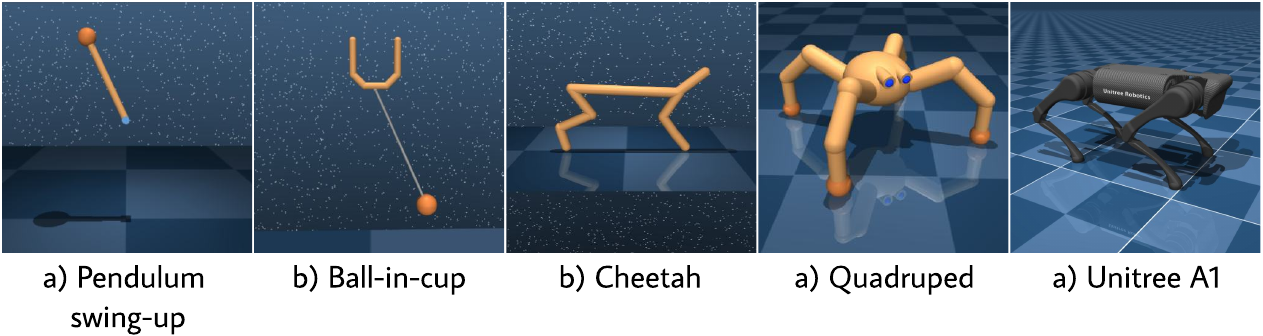}
    \caption{Visualizations of the systems used as benchmarking tasks. Each task's state and action dimensions are reported in parentheses as S and A respectively.}
    \label{fig:tasks}
    \vspace{-0.3cm}
\end{figure}

We test our framework on a set of standardized control tasks of increasing difficulty and spanning a wide range of dimensions of state and input spaces. We applied our method to find programmatic policies for the \textit{pendulum swing-up}, \textit{ball-in-cup} tasks, as well as three locomotion tasks: \textit{dm\_control cheetah}, \textit{dm\_control quadruped}, and \textit{Unitree A1} (see Fig.~\ref{fig:tasks}). The respective dimensions of state and input spaces are reported in Tab. \ref{tab:task-dimensions}.  
%
\paragraph*{Pendulum swing-up} This task consists of controlling an input-constrained pendulum to reach the upright equilibrium. The maximum applicable torque is $1/6^{\text{th}}$ as required to lift it from a horizontal motionless position. Even if low-dimensional, this task is not easily solvable. The pendulum has to oscillate and accumulate enough energy to be able to swing to the upright configuration.
For this task, the step reward \eqref{eq:step-reward} is defined as 
\begin{equation}
    r_t = \begin{cases}
  1& \text{if} \,\, |\theta| < 0.5\\
  0 & \text{otherwise}
\end{cases}
\end{equation}
where $\theta\in[-\pi, \pi)$ is the angle with respect to the upward equilibrium configuration.
\paragraph*{Ball-in-Cup}
The \textit{ball-in-cup} system is composed of a two-dimensional double integrator (the cup) and a ball attached to it through a string (a unilateral distance constraint between the two objects). The system is planar. The task consists of finding a policy for the cup to catch the ball. In this case, the policy outputs reference horizontal and vertical positions, which are tracked by a lower-level linear controller.
For this task, the step reward \eqref{eq:step-reward} is defined as 
\begin{equation}
    r_t = \begin{cases}
  1& \text{if} \,\, \text{ball inside cup}\\
  0 & \text{otherwise.}
\end{cases}
\end{equation}
\paragraph*{Cheetah} This is the simplest of the locomotion tasks. The task consists of controlling a planar biped to move forward by outputting joint angles, also tracked by lower-level linear controllers. The step reward \eqref{eq:step-reward} is defined as 
\begin{equation}
    r_t = \max(0,\, \min(1,\, v/10)),\label{eq:locomotion-rew}
\end{equation}
where $v$ is the horizontal velocity of the biped torso.
\paragraph*{Quadruped} This is a three-dimensional locomotion task. The system to control is a four-legged robot, with three actuated joints per leg. The step reward function is the same as \eqref{eq:locomotion-rew}. The goal is to make the quadruped move forward.
\paragraph*{Unitree A1} We also tested our framework to generate a gait for a simulated Unitree A1 robot. We used the official model from the MuJoCo Menagerie \cite{menagerie2022github}. The Unitree A1 is one of the most widely used quadrupeds in robotics research, and is therefore the most realistic of the locomotion tasks. In this case, we decided to generate a gait that would make the robot walk backward. The locomotion reward is the same as~\eqref{eq:locomotion-rew}, using the appropriate direction and sign for the velocity.

When simulated, the dynamics equations are discretized through a semi-implicit Euler method (details in \cite{todorov2012mujoco}). The controls are injected every $20 \, \mathrm{ms}$ for the \textit{pendulum swing-up}, \textit{ball-in-cup} and \textit{quadruped} tasks, every $10 \, \mathrm{ms}$ for the \textit{cheetah}, and every $1 \, \mathrm{ms}$ for the \textit{Unitree A1}. All episodes are $1000$ control steps long. In all cases, our approach:
\begin{itemize}[leftmargin=*]
    \item Shows higher performance compared to optimization-free program search, in some cases by margins of $100\%$ or~more;
    \item Automatically finds high performing static feedback policies in the form of \eqref{eq:param-prog} that are lightweight and interpretable.
\end{itemize}
\begin{figure*}[tb] 
    \centering
    \includegraphics[width=1.0\textwidth]{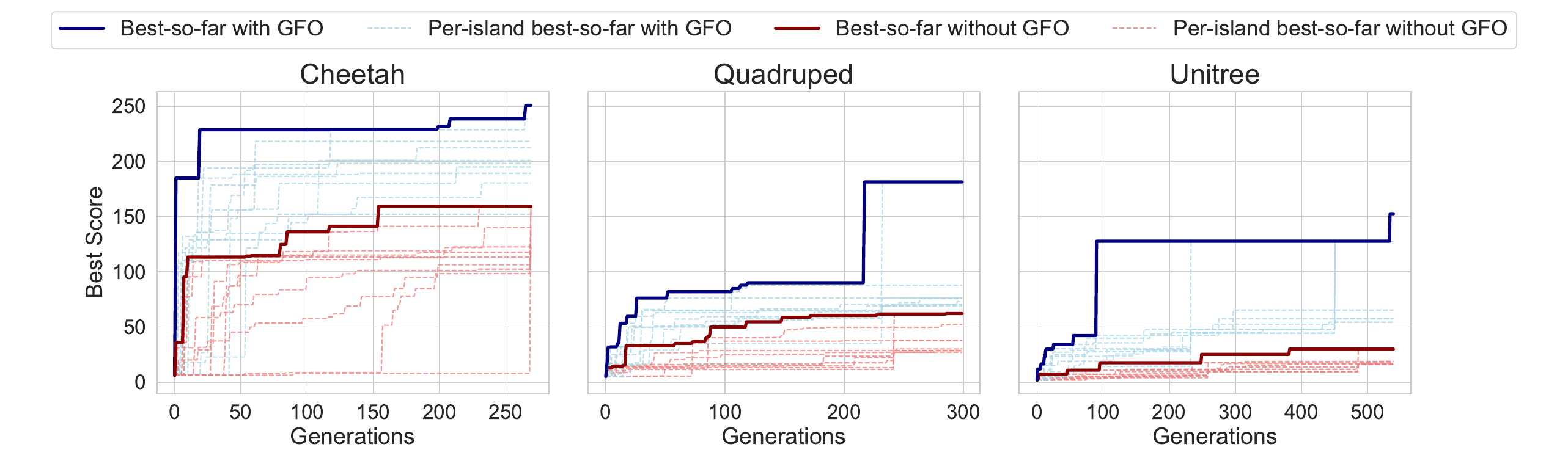} 
    \caption{Best-so-far curves of policy search runs with in-the-loop GFO and without, respectively in blue and red. The solid lines represent the best-so-far score across all islands, while the dashed lines represent the best-so-far score within each island.}
    \label{fig:running-best}
    \vspace{-0.3cm}
\end{figure*}

We report best-so-far curves for the locomotion tasks in Fig.~\ref{fig:running-best}, and a quantitative comparison of the rewards achieved by the policies found for all tasks with in-the-loop GFO and without in Tab.~\ref{tab:task-dimensions}. For each task
The performances in Tab.~\ref{tab:task-dimensions} for the \textit{pendulum swing-up} and \textit{ball-in-cup} tasks are averaged across 1000 episodes in which the initial conditions are randomized. In fact, these two tasks are more sensitive to initial conditions than the locomotion tasks. For \textit{Cheetah}, \textit{Quadruped}, and \textit{Unitree A1}, the performance is averaged across 10 episodes, which is the same number of episodes used to compute the programs' score during the synthesis process. For these tasks, the policies achieve average forward torso velocities of $2.5\,\mathrm{m\cdot s^{-1}}$, $1.8\,\mathrm{m\cdot s^{-1}}$, and $-1.52\,\mathrm{m \cdot s^{-1}}$ (backward walk), respectively. 

With reference to Fig. \ref{fig:running-best}, all the runs were carried out with a fixed wall-clock time budget: $2$ hours for \textit{pendulum swing-up} and \textit{ball-in-cup}, $5$ hours for \textit{cheetah} and \textit{quadruped}, and $10$ hours for \textit{Unitree A1}. Had we allowed for more wall-clock time, even better policies would probably have been found by the search process. One more way to interpret the best-so-far curves is in a sample efficiency sense: taking a target score, i.e. a horizontal line in the graphs, how many generations will be needed approximately to achieve it. It is then clear that introducing the GFO loop greatly improves the sample efficiency of the search process.
It is also interesting to note that the GFO seems to provide greater benefit for higher-dimensional state and action spaces, thus bringing these methods closer to be deployed in real-world systems. 

We report the best policies found for the five tasks in Fig.~\ref{fig:codesnippets}. In all cases, the found policies are below 35 lines of code, and can be read, edited, ad debugged. In all cases except \textit{quadruped}, the policies contain fewer than 10 parameters. The comments and variable names in the snippets are produced by the LLM. By identifying what the variables represent and their bounds, the policy could be simplified even more. The shared language (code) allows a user to iterate in the loop with our synthesis algorithm as well.
Videos showing the systems' behaviors with the respective policies running in closed-loop can be found at \url{https://sites.google.com/berkeley.edu/colmo}.
%

    
%
\begin{figure*}[b] 
  \centering

  \begin{minipage}[t]{0.49\textwidth}
  \begin{subfigure}[t]{\textwidth}
    \begin{lstlisting}[language=Python, escapechar=!, basicstyle=\ttfamily\scriptsize\fbseries]
def policy(obs):
  """obs size is 3. return shape is ()"""
  x1 = np.arctan2(-obs[1], obs[0]) 
  # Adjust action
  if np.abs(x1) > np.pi / !\hlcode{6.512}!:
    action = np.sign(obs[2])
  else:
    action = !\hlcode{9.437}!*np.sin(x1) - !\hlcode{2.746}! * obs[2]

  return action
\end{lstlisting}
\caption{Pendulum swing-up.}
  \end{subfigure}

    \begin{subfigure}[t]{\textwidth}
    \begin{lstlisting}[language=Python, escapechar=!, basicstyle=\ttfamily\scriptsize\fbseries]
def policy(obs):
  """obs size is 17: position (8), velocity (9). 
  return shape is (6,)."""

  pos = obs[:3]
  vel = obs[3:6]
  ang_vel = obs[6:9]

  # Adjust forward control gain
  action = np.zeros(6)
  action[0] = !\hlcode{-12.241}! * vel[0]  

  action[1] = !\hlcode{-0.603}! * ang_vel[0]  # Hip angle
  action[2] = !\hlcode{11.351}! * ang_vel[1]  # Knee angle
  action[3] = !\hlcode{-8.106}! * ang_vel[2]  # Ankle angle

  # Introduce a new control term for balance
  action[4] = !\hlcode{-12.268}! * ang_vel[1]

  return action
\end{lstlisting}
    \caption{Cheetah}
  \end{subfigure}

  \begin{subfigure}[t]{\textwidth}
    \begin{lstlisting}[language=Python, escapechar=!, basicstyle=\ttfamily\scriptsize\fbseries]
def policy(obs):
  torso_upright = obs[47]
  imu = obs[48:54]
    
  if torso_upright < !\hlcode{1.131}! or any(abs(x) > !\hlcode{0.529}! for x in imu):
    # Adjust actions for better stability
    actions = np.array([!\hlcode{-0.4, -0.499, 1.321}!, !\hlcode{-1.483, 0.774, -0.281, 1.049, -0.836, -0.195}!, !\hlcode{0.846, 1.898, 2.066}!])
  elif abs(torso_upright) < !\hlcode{0.825}!:
    # Adjust actions for a balanced stance
    actions = np.array([!\hlcode{0.2, -0.493, -0.287, 0.861, -2.279, 0.03}!, !\hlcode{-1.584, -0.079, 0.427, 0.233, 0.801, 0.163}!])
  else:
    actions = np.array([!\hlcode{0.0, -0.173, -0.024}!, !\hlcode{2.506, -0.82, 0.029, -1.302, 0.009, -1.243}!, !\hlcode{0.057, -1.264, -1.961]}!)
    
  if obs[-1] < !\hlcode{-0.275}!: # Performance adjustment
    actions *= !\hlcode{-1.29}!
  elif obs[-1] > !\hlcode{1.442}!:
    actions *= !\hlcode{-1.893}!
    
  # Minimal noise addition for fine-tuning
  actions += np.random.normal(scale=0.04, size=12)
    
  return actions
\end{lstlisting}
    \caption{Quadruped}
  \end{subfigure}

\end{minipage}
\hfill
\begin{minipage}[t]{0.49\textwidth}
  \begin{subfigure}[t]{\textwidth}
    \begin{lstlisting}[language=Python, escapechar=!, basicstyle=\ttfamily\scriptsize\fbseries]
def policy(obs)                         
  """obs size is 8. return shape is (2,)."""

  ball_pos = obs[:3] 
  cup_pos = obs[3:]   
  action = np.zeros((2,))                

  # Further refine horizontal movement
  if abs(ball_pos[0] - cup_pos[0]) > !\hlcode{-12.071}!:
    action[0] += (cup_pos[0] - ball_pos[0]) * !\hlcode{-5.66}!

  # Fine-tune vertical position correction
  if abs(ball_pos[2] - cup_pos[2]) > !\hlcode{-7.763}!:
    action[1] += (cup_pos[2] - ball_pos[2]) * !\hlcode{1.023}!

  # Add slight bias to vertical movement
  action[1] += !\hlcode{0.531}!

  return action
    \end{lstlisting}
    \caption{Ball-in-cup}
  \end{subfigure}
\vspace{0.3cm}

  \begin{subfigure}{\textwidth}
    \begin{lstlisting}[language=Python, escapechar=!, basicstyle=\ttfamily\scriptsize\fbseries]
def policy(obs):
  """obs size is 27: position and angles (19), velocities (8). return shape is (12,)"""

  joint_angles = obs[0:12]
  joint_velocities = obs[12:24]
  body_velocity = obs[24:27]

  target_joint_angles = joint_angles.copy()

  # Adjust the hip joints
  target_joint_angles[0::3] += !\hlcode{-27.697}! * np.sign(body_velocity[0])
  target_joint_angles[1::3] += !\hlcode{-6.726}! * np.sign(body_velocity[0])

  # Adjust the knee joints
  target_joint_angles[1::3] += !\hlcode{6.09}! * joint_velocities[1::3]

  # Adjusting the shoulder joints
  target_joint_angles[2::3] += !\hlcode{-14.423}!

  # Add a simple spring-like mechanism
  balance_factor = !\hlcode{17.69}!
  target_joint_angles[0::3] += balance_factor * (body_velocity[1] + body_velocity[2])
  target_joint_angles[1::3] += balance_factor * (body_velocity[1] - body_velocity[2])

  # Add additional balancing adjustments
  if body_velocity[2] < !\hlcode{-32.662}!:
    target_joint_angles[0::3] += !\hlcode{10.462}!
    target_joint_angles[1::3] += !\hlcode{-22.429}!

  # Adjustments to ensure proper leg movement
  target_joint_angles[0::3] += !\hlcode{14.31}!
  target_joint_angles[1::3] -= !\hlcode{-49.337}!

  # Return the target joint angles
  return target_joint_angles
\end{lstlisting}
    \caption{Unitree A1}
  \end{subfigure}
\end{minipage}
  \caption{Policies found by our synthesis algorithm for the control tasks considered. The parameters optimized by the GFO loop are highlighted in orange.}
  \label{fig:codesnippets}
\end{figure*}
\section{Conclusions \label{sec:conclusion}}
%
We presented a hybrid symbolic-numeric approach to automatically synthesize interpretable control policies combining LLM-guided program search with in-the-loop gradient-free optimization. By decoupling the synthesis of a control program from the selection of its parameters, our method leverages the complementary strengths of symbolic representations and numerical optimization methods.

Across a set of control benchmarks, we demonstrate that our approach achieves higher performance and faster convergence than purely LLM-driven search while maintaining the interpretability and modularity of the resulting policies. The policies we find are also simpler and carry orders of magnitude fewer parameters than neural network policies solving the same tasks through reinforcement learning. 
This compactness in terms of parameter count not only improves interpretability but also facilitates deployment on resource-constrained systems where memory and compute efficiency are critical.
These results suggest that jointly exploring symbolic and numeric spaces leads to reliable and efficient discovery of control strategies. 

By demonstrating promising results without resorting to prohibitively expensive computational resources, we showed the accessibility and practicality of our approach, which brings us closer to fast programmatic control design iterations and more efficient turnaround times when developing control policies. We believe our method is an important step toward automatically synthesizing interpretable, verifiable, and high-performing controllers for real-world systems.

This work opens several promising directions. Future research could extend this approach to more complex policy representations, such as time-varying functions or policies with memory (e.g. access not only to the current observation, but also to past history) as well as differentiable code generation and gradient-based in-the-loop optimization. 
\section*{ACKNOWLEDGMENT}
This work has been partially supported by the Cal-Next Solar Center, \url{https://cal-next.berkeley.edu/} and the UC Berkeley College of Engineering. The authors would like to thank Orr Paradise for the insightful conversations and the members of HiPeRLab (\url{https://hiperlab.berkeley.edu/members/}) and SCALE Research Lab (\url{https://scale.berkeley.edu/people/}) for their feedback. 
%
%
%
\bibliography{IEEEabrv, refs}
\end{document}